\documentclass[balance,upint,subscriptcorrection,varvw,mathalfa=cal=boondoxo,spanish,french,vietnamese,russian,greek,pdf-a,fontspec,colorlinks,nofoot]{asmeconf}

\usepackage{packages}

\usepackage{url}

\usepackage{breakurl}

\hypersetup{%
	pdfauthor={Florian Stadtmann},
	pdftitle={Anomaly Detection in Floating Offshore Wind Energy through Diagnostic Digital Twin},                  
	pdfkeywords={Public informed opinion, digital twin, decision making},
}

\begin{document}

	
	\title{Diagnostic Digital Twin for Anomaly Detection in Floating Offshore Wind Energy} 
	
	%
	%
	%
	
	\SetAuthors{%
		Florian Stadtmann\affil{1}\CorrespondingAuthor{},
		Adil Rasheed\affil{1}\affil{2}
	}
	
	\SetAffiliation{1}{Norwegian University of Science and Technology, Trondheim, Norway}
        \SetAffiliation{2}{Department of Mathematics and Cybernetics, SINTEF Digital, Trondheim, Norway}
	
	
	\maketitle
	\begingroup
	\renewcommand
	\thefootnote{\textsection}
	\endgroup
	
	
	
	\normalfont
	

	\begin{abstract} 
        The demand for condition-based and predictive maintenance is rising across industries, especially for remote, high-value, and high-risk assets. In this article, the diagnostic digital twin concept is introduced, discussed, and implemented for a floating offshore turbine.
        A diagnostic digital twin is a virtual representation of an asset that combines real-time data and models to monitor damage, detect anomalies, and diagnose failures, thereby enabling condition-based and predictive maintenance. By applying diagnostic digital twins to offshore assets, unexpected failures can be alleviated, but the implementation can prove challenging. Here, a diagnostic digital twin is implemented for an operational floating offshore wind turbine. The asset is monitored through measurements. Unsupervised learning methods are employed to build a normal operation model, detect anomalies, and provide a fault diagnosis. Warnings and diagnoses are sent through text messages, and a more detailed diagnosis can be accessed in a virtual reality interface. The diagnostic digital twin successfully detected an anomaly with high confidence hours before a failure occurred. 
        The paper concludes by discussing diagnostic digital twins in the broader context of offshore engineering. The presented approach can be generalized to other offshore assets to improve maintenance and increase the lifetime, efficiency, and sustainability of offshore assets.

        \end{abstract}
    
    
    \section{Introduction}
    Wind energy is increasingly moving offshore, and the potential for floating offshore wind in deep waters is substantial~\cite{2018ida,Freeman2019oeo,IEA2019owo}. However, these remote wind farms are more difficult to access and inspect ~\cite{Adumene2022oss}. Furthermore, the response time to failures can be much higher since weather conditions and the availability of technicians, service vessels, and spare parts need to be balanced~\cite{Irawan2017oom}. Unexpected failures can lead to months of turbine downtime thereby reducing both revenue and energy efficiency of the turbine. To avoid unexpected downtime, current preventive maintenance strategies balance energy production and maintenance costs based on average component lifetimes~\cite{Ren2021owt}.  
    \begin{figure*}[ht]
        \centering
        \begin{subfigure}[t]{0.71\linewidth}
            \includegraphics[width=\linewidth]{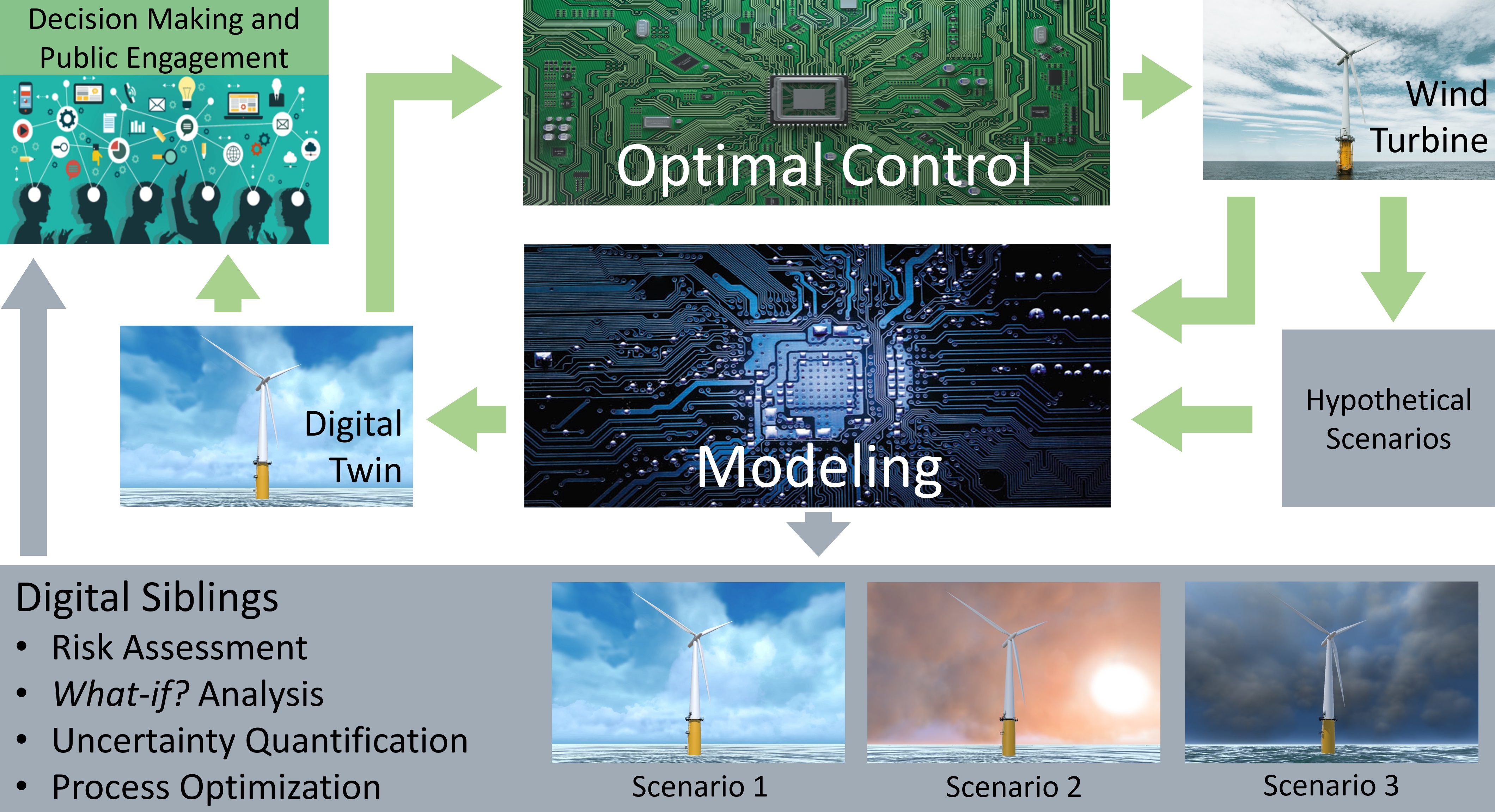}
            \caption{Digital twin framework }
            \label{fig:dt}
        \end{subfigure}\hfill
        \begin{subfigure}[t]{0.27\linewidth}
            \includegraphics[width=\linewidth]{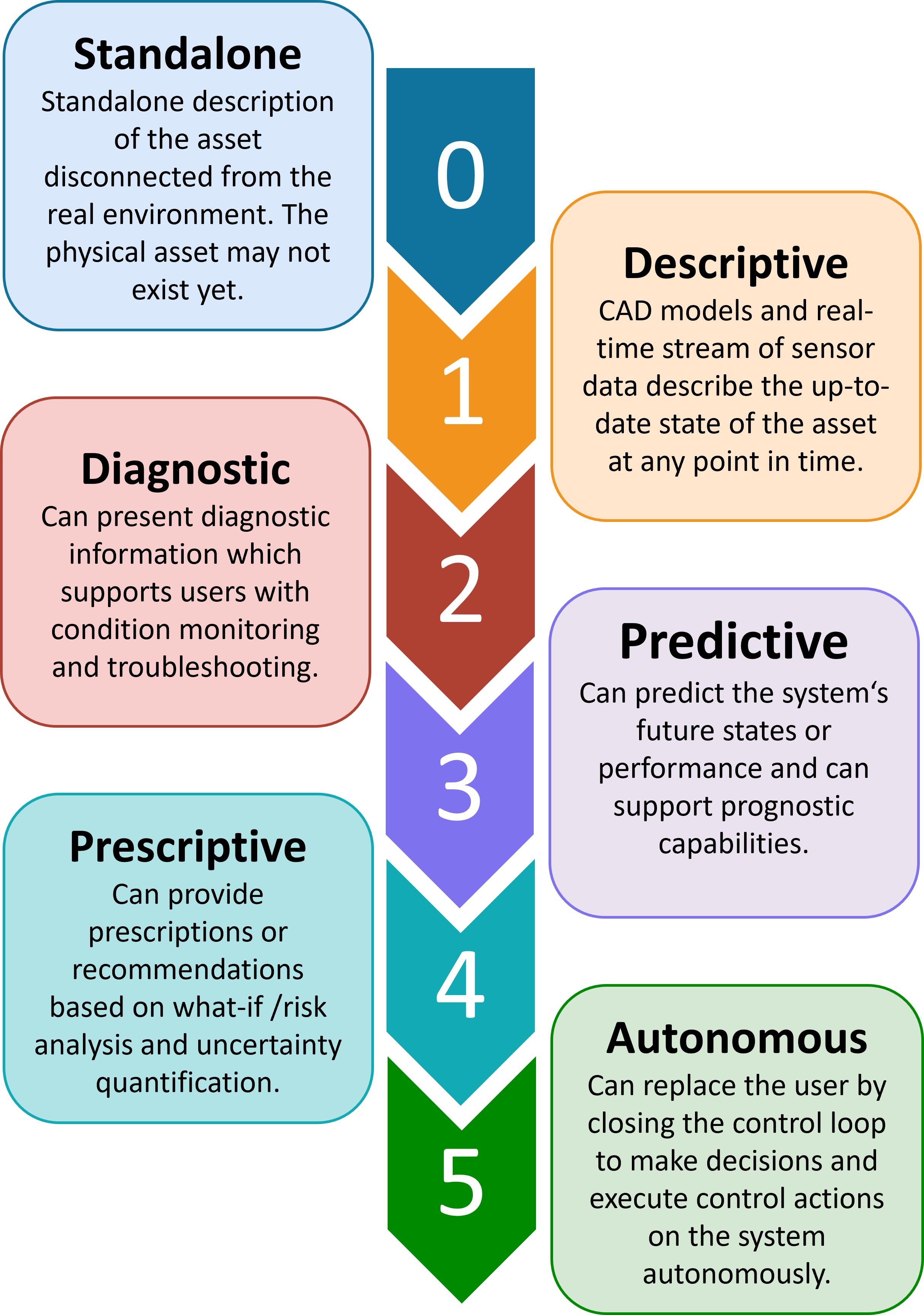}
            \caption{Capability level scale}
            \label{fig:dt_cl}
        \end{subfigure}
        \caption{Digital twin framework and capability level scale~\cite{Stadtmann2023doa}.}
        \label{fig:dt_both}
    \end{figure*}
    On one hand, these maintenance strategies result in the replacement of components long before their lifetimes expire and ultimately lead to the decommissioning of turbines that could be operated for several more years. On the other hand, unexpected failures and therefore unexpected downtime still occur. If the remaining useful lifetime of components could be estimated based on the actual locally experienced conditions and formerly unexpected failures could be predicted from small anomalies, the maintenance strategy could be changed from reactive and preventive to condition-based and predictive.
    It is therefore not surprising that there is a large industrial interest in condition monitoring, condition-based maintenance, and predictive maintenance, especially by offshore industries with remote assets.
    Digital twins are a key enabler in facilitating this shift in maintenance strategies by monitoring and analyzing the asset condition.
    
    In this work, the concept of diagnostic digital twins is presented, discussed, and implemented for a floating wind turbine. The aim of this article is 
    \begin{itemize}
        \item to familiarize the reader with the concept of digital twins in general and diagnostic digital twins in particular.
        \item to present challenges and enablers connected to diagnostic digital twins.
        \item to demonstrate the application of diagnostic digital twins by integrating them for a floating offshore wind turbine.
    \end{itemize}
    
    In Section~\ref{sec:Theory} the concept of digital twins, capability levels, and diagnostic digital twins is presented, and relevant methods employed in this work are explained. Section~\ref{sec:Method} addresses the integration of diagnostic features into the digital twin. In Section~\ref{sec:Results and Discussion} the results from evaluating 
    the diagnostic features on data from an existing floating wind turbine are presented. Moreover, their adaptation to other areas of offshore engineering is discussed. The work is concluded in Section~\ref{sec:Conclusion}, and recommendations for future work are provided.
    
    \section{Theory} 
    \label{sec:Theory}
    Here a brief introduction to digital twins and capability levels is given. It follows a short overview of some relevant work on anomaly detection and an explanation of the data-driven models used. 
        \subsection{Digital Twins and Capability Levels}
            Digital twins are virtual representations of a physical asset or system enabled through data and simulators for real-time prediction, optimization, monitoring, control, and informed decision-making \cite{Rasheed2020dtv,Stadtmann2023dti}. The digital twin framework is shown in Figure~\ref{fig:dt}.
            A digital twin (left) of a physical asset (top right) mirrors the asset in appearance, state, and behavior at any time. It is synchronized with the asset through real-time measurements. However, since measurements are limited in spatial and temporal resolution, models (middle) are needed to interpolate and augment the data. A single digital twin can fulfill many objectives such as communicating concepts, aggregating and analyzing data, visualizing data insights, predicting future asset state and behavior, providing decision support, and closing the loop through autonomous asset control.
            The broad variety of applications and purposes for digital twins can easily cause misunderstandings about the digital twin concept since the capabilities required in (and therefore integrated into) a digital twin depend on the purpose.
            A classification scheme based on capabilities is required to clearly distinguish between digital twins with different purposes and therefore capabilities. The capability level scale~\cite{Sundby2021gcd, DNVGL2020dra,Elfarri2023aid,Stadtmann2023dti} groups digital twins into six levels (level 0-5) as shown in Figure~\ref{fig:dt_cl}. The standalone digital twin is a virtual representation of an asset that is not synchronized with the physical asset. While it may be used for design, planning, visualization, communicating, and outreach, it lacks the real-time connection to the asset and is therefore classified as level 0. Once a real-time connection from the asset to the virtual representation is established, the digital twin is labeled as descriptive (level 1) - it describes the current and past asset states and can be used for data aggregation and visualization. The descriptive digital twin provides the basis for higher capability levels as it is an aggregate of data with a suitable interface to visualize the results of analyses performed in higher capability levels.
            The data collected in the descriptive digital twin can be used to track and analyze the current and past asset conditions and detect anomalies in the asset. A digital twin equipped with such capabilities is called a diagnostic digital twin (level 2). So far the digital twin has used the current and past asset state. Utilizing the future asset states requires predictive algorithms, and digital twins incorporating such features are labeled predictive (level 3). Diagnostic and predictive capabilities can be combined with uncertainty quantifications, what-if analyses, and risk estimation to recommend actions to a user, at which point the digital twin is called prescriptive (level 4). Finally, the digital twin may influence the asset directly without a human user in the loop, at which point it is classified as autonomous (level 5). A more detailed explanation of the capability level scales can be found for example in~\cite{Stadtmann2023dti,Stadtmann2023doa}.
            This work focuses on the diagnostic digital twin and specifically anomaly detection capabilities. For the integration of standalone, descriptive, and predictive digital twins, the authors refer to their earlier work in~\cite{Stadtmann2023doa,Stadtmann2023sda}.

        \subsection{Anomaly Detection}
            The diagnostic digital twin uses the current and past asset state to provide information on the condition of the asset and its components in real-time.
            There exist various condition-monitoring strategies. Some track cumulative damage, as is typical for example in structural health monitoring through stress measurements, Rainflow method, stress-life curves, and Miner's rule~\cite{Pacheco2022fao}. Such methods typically give insights into the component's health but require information about the material and design of the component. Anomaly detection methods monitor signals to detect deviations from the normal asset operation.
            To detect an anomaly it is sufficient to build a model that describes the normal operation - a normal operation model (NOM). Anything that cannot be explained by the NOM is classified as an anomaly.
            Anomaly detection algorithms are strongly leaning towards data-driven methods, especially neural-network-based approaches such as feed-forward neural networks \cite{Sun2016agm}, nonlinear autoregressive exogenous models \cite{Cui2018aad, McKinnon2020eot}, autoencoders \cite{Roelofs2021aba,Zhang2022ada,Renstrom2020swa}, long short-term memory neural networks (LSTM) \cite{Vidal2023pmo}, and combinations of methods~\cite{Chen2021ada,Zhang2024roa,Zhang2023adm,Zhu2022ada}. A strong advantage of data-driven models as a NOM is that training them requires little to no domain knowledge and only data from normal operation, not from failures, which makes it possible to use unsupervised learning methods.
            In this work, dense neural networks (DNNs) and LSTMs are used as NOMs.

        \subsection{Dense Neural Networks}
\begin{figure}
            \centering
            \begin{subfigure}{\linewidth}
            \centering
            \includegraphics[width=0.8\linewidth,trim= 0 80 0 0,clip]{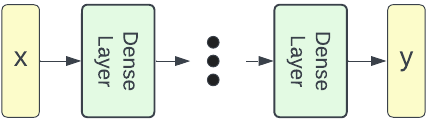}
            \caption{Dense neural network structure}
            \label{fig:DNNstruc}
            \end{subfigure}
            \begin{subfigure}{\linewidth}
            \centering
            \includegraphics[width=0.8\linewidth,trim= 0 0 0 60,clip]{Figures/Dense_NN.pdf}
            \caption{Single dense layer}
            \label{fig:dnn_layer}
            \end{subfigure}
            \caption{Dense neural network structure}
            \label{fig:enter-label}
        \end{figure}

            Dense neural networks (DNNs), also referred to as fully connected neural networks, are among the most used types of neural networks. They consist of layers, where each layer takes an input vector $x$ from the previous layer and returns an output vector equal to the number of nodes in the layer. The output is calculated with
            \begin{align}
                y &= \mathrm{A}\left(W x + b\right)
            \end{align}
            where $W$ is a matrix of weights and $b$ a vector of biases. Each layer has its own weights and biases. $\mathrm{A}()$ is called the activation function and is applied element-wise. Typical activation functions include hyperbolic tangent, rectified linear units (ReLU), and 
            leaky rectified linear units. Here, the ReLU activation function is used, which is defined as
            \begin{align}
                \mathrm{ReLU}(x) &= max(0,x)
            \end{align}  
            Figure~\ref{fig:DNNstruc} shows an example of the layer structure, and the operation in a layer is depicted in Figure~\ref{fig:dnn_layer}.
            
        \subsection{Long Short-Term Memory Neural Networks}
\begin{figure}[t]
            \centering
            \begin{subfigure}{\linewidth}
            \centering
            \includegraphics[width=\linewidth,trim= 0 10 0 0,clip]{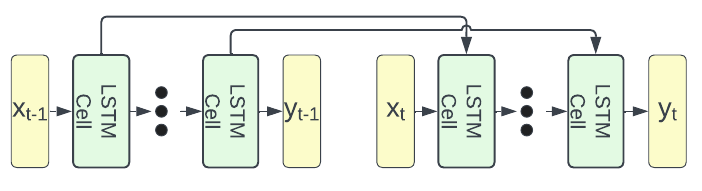}
            \caption{Long short-term memory neural network (LSTM) structure}
            \label{fig:LSTMstruc}
            \end{subfigure}
            \begin{subfigure}{\linewidth}
            \centering
            \includegraphics[width=\linewidth]{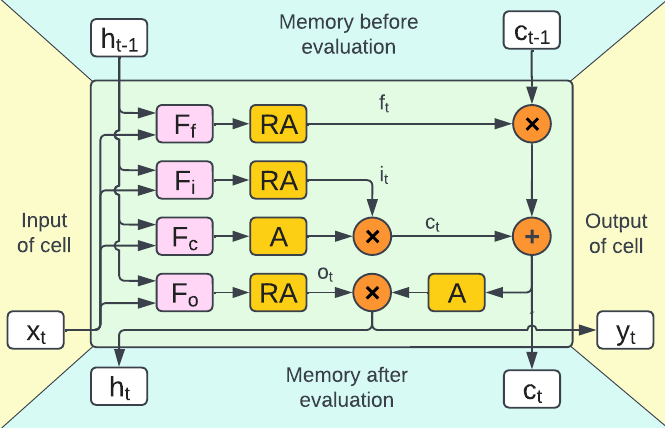}
            \caption{Single long short-term memory layer}
            \label{fig:LSTM_Cell}
            \end{subfigure}
            \begin{subfigure}{\linewidth}
            \centering
            \includegraphics[width=\linewidth,trim= 0 5 0 0,clip]{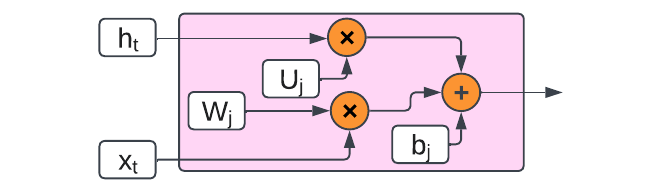}
            \caption{$F_j$ operation}
            \label{fig:LSTMFunc}
            \end{subfigure}
            \caption{Long short-term memory neural network structure}
            \label{fig:LSTM}
        \end{figure}

            Long short-term memory neural networks (LSTMs)~\cite{Hochreiter1997lst,Gers1999ltf} are a version of recurrent neural networks that can memorize information throughout long data series. This is made possible by including a memory component inside each node in the neural network. The output of an LSTM node during the evaluation of one sample is used to update a hidden state and a cell state in each LSTM node, which is carried over between consecutive evaluations. The calculation of the hidden state and cell state of a node is done with three gates, the forget gate $f$, the input gate $i$, and the output gate $o$. The forget gate $f$ and input gate $i$ determine how much of the previous state $c_t$ and input $i_t$ goes into the next state $c_{t+1}$ respectively. The output gate $o$ determines how much of the current state $c_t$ is being sent to the following nodes. The parameters are calculated with
            \begin{align}
                   f_t &= \mathrm{RA}\left(W_f x_t + U_f h_{t-1} + b_f\right) \\
                   i_t &= \mathrm{RA}\left(W_i x_t + U_i h_{t-1} + b_i\right) \\
                   o_t &= \mathrm{RA}\left(W_o x_t + U_o h_{t-1} + b_o\right) \\
                   c_t &= f_t \odot c_{t-1} + i_t \odot \mathrm{A}\left( W_c x_t  + U_c h_{t-1} + b_c \right) \\
                   h_t &= o_t \odot \mathrm{A}\left(c_t\right)
            \end{align}
            where $\mathrm{A}()$ is the activation function, $\mathrm{RA}()$ is the recurrent activation, and $\odot$ is the elementwise product. $W_j$ and $U_j$ are the metrics of each gate $j$ that contain the weights of the input and hidden state respectively. $b_j$ are the bias vectors of each gate.
            The structure of an LSTM is shown in Figure~\ref{fig:LSTM}.

    \section{Methodology} \label{sec:Method}
        With the theory in place, this section presents the data used, explains the selection features, and describes the setup of the NOM models integrated in this work. Following this, the anomaly detection algorithm is introduced, and the approach for anomaly diagnosis is presented. The section concludes by briefly outlining the integration into the digital twin.
        \subsection{Data}
            This study uses supervisory control and data acquisition (SCADA) data measured at the operational floating offshore wind turbine Zefyros. The Norwegian research turbine was the first full-scale floating offshore wind turbine \cite{Driscoll2016voa}. Formerly known as Hywind Demo, it was built in 2008 and has been the prototype turbine for the floating wind farms Hywind Scotland and Hywind Tampen \cite{Ibrion2023oar}. The turbine uses the Siemens SWT-2.3MW-82 mounted on a steel spar buoy floater~\cite{Skaare2015aom}.
            One year of SCADA data from the floating wind turbine is used. The raw data set consists of several signals, including but not limited to wind speed, active and reactive power, rotor and generator revolutions per minute (RPM), and temperature measurements of the shaft bearing, shaft breaks, generator stator, generator rotor, and gearbox oil. The measurement frequency varies, but it averages at 0.3~Hz. However, such high measurement frequencies are often not available. Here, the data is averaged to one-minute intervals resulting in around 525.000 samples.
            The data are complemented by the status code and operational code of the turbine. The codes can be used to identify the start and duration of faults, but it does not give any information on the source of a fault.

        \subsection{Feature Selection}
            The goal of the descriptive digital twin implemented here is to identify anomalies in the data in real-time that may result in faults. Anomaly detection by vibration and acoustic emission requires high measurement frequencies and data is not available here. However, the measured temperature of components may capture abnormal behaviour, and much lower measurement frequencies are required due to the relatively slow change of temperature signals.
            Therefore, all temperature signals measured in the gearbox are used as the endogenous variables, namely the temperatures of the oil, the shaft bearing, the generator rotor, the generator stator, and two measurements of the shaft brakes. Furthermore, the environment temperature and the RPM of the rotor are added as exogenous variables in all models to provide information on the operating conditions and the environment. The wind speed, active power, and generator RPM have been excluded as part of the feature selection since they are strongly correlated with the rotor RPM.

        \subsection{Normal Operation Model}   
            \label{ssec:m_nom}
        
            In this work, NOMs are built with DNNs and LSTMs. The DNNs use the above-mentioned eight variables as input and a single variable as the output. Each of the endogenous variables is predicted with a separate DNN NOM. The motivation for this choice is that a single-output model is optimized toward the prediction of each single parameter, while a multi-output model has to compromise between parameter accuracies. A comparison with multi-output models has shown that the single-output models achieve better overall performance.
            Note that the model inputs and outputs have no time shift, which makes the model robust against missing values and non-equidistant measurement intervals.
            The DNN architecture, learning rate, and batch size have been manually tuned. It was observed that small changes to the architecture and hyperparameters do not impact the model performance. Notably larger networks tend to overtrain, while smaller neural networks are not able to learn the normal operating conditions.
            It should be noted that the neural networks can only be optimized toward learning the normal operating conditions, not towards detecting anomalies since the training is unsupervised.
            The tuned DNN architecture consists of three layers with eight, five, and one node respectively, where all layers use ReLU activation except for the last one. The Adam optimizer is used for training with a learning rate of 0.01 and batch size of 32. The loss is chosen as mean absolute error to reduce the training sensibility to outliers and anomalies.
            The model is trained for only 5 epochs since fast convergence is observed on the 315,000 training samples. However, neural networks are known to converge to local minima which may vary when training the same model multiple times with different initial weights and shuffled data. Therefore, the model is retrained three times, and the best model is chosen, which makes the resulting model more robust towards retraining.
            \begin{figure}[t]
                \centering
                \includegraphics[width=\linewidth]{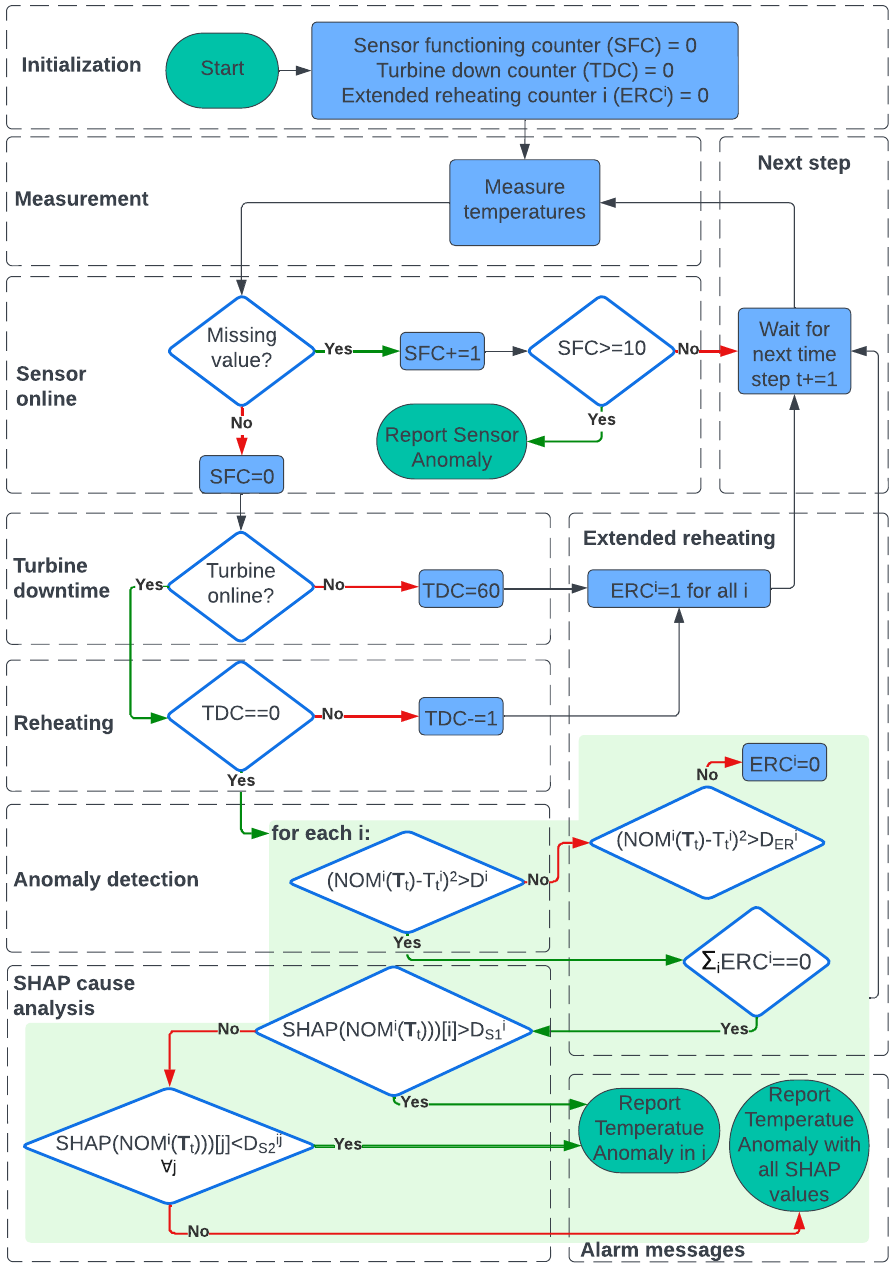}
                \caption{Anomaly detection algorithm using a normal operation model and taking into account missing sensor values, turbine downtime, reheating process, and extended reheating. For more details see Section~\ref{ssec:m_anom}.
                }
                \label{fig:anom}
            \end{figure}
            It should be noted that the DNN target value is already contained in the input, and so a perfect DNN would simply return that value. However, the DNN does not learn this behavior due to correlations between the parameters. The fact that the DNN consistently does not learn to return the input value but rather bases its estimate on certain parameter combinations is used as an advantage here to identify anomalies.
            
            The data set is split into 60\% training and 40\% testing data. The split is chosen so that the few longer downtimes that are initiated by a fault are located in the test set and can be used for validation.
            The inputs of the model are normalized on the training data to [0,1].
            Known downtimes are excluded from the data during training. A turbine downtime will show in the data by a gradual cooling of all components, followed by a reheating of components once the turbine is operational again. An additional hour after each known turbine downtime is excluded during training to allow for this reheating to take place.
            A second set of NOMs is built with LSTMs. 
            In contrast to the DNNs, the LSTMs are tasked to predict the temperature measurements one step ahead.
            The memory of the LSTM makes it inherently time-aware. On one hand, it opens the potential for detecting unusual time sequences in the data. On the other hand, it is sensible to missing values. Furthermore, the model can be corrupted by anomalies even after they have passed since the hidden states and cell states of the LSTM can be overwritten by large anomalies not present in the training data.
            The network architecture and hyperparameters underwent another manual tuning which showed that the LSTMs are performing well with a similar architecture, learning rate, and batch size as the DNNs. It is trained on a time series of 30 minutes length each. Larger values do not improve the training and increase the training cost, but significantly smaller values may prevent the LSTM from learning long-term dependencies. Note again that the neural networks can only be optimized toward learning the normal operating conditions, not towards detecting anomalies since the training is unsupervised.
            
        \subsection{Anomaly detection}
        \label{ssec:m_anom}
            An anomaly is identified if the mean squared error (MSE) between the NOM prediction for a temperature variable $T_t^i$ at time $t$ and the measured value is larger than a set threshold $D^i$.
            \begin{align}
                \mathrm{Anomaly}_t^i = \begin{cases}
                    1,&  \left(\mathrm{NOM}^i\left(\bm{T}_t\right)-T^i_t\right)^2 > D^i \\
                    0,& else
                \end{cases} 
            \end{align}
            The threshold $D^i$ is set to be larger than the maximum deviation between NOM and measurement at all times without fault $F$, close enough before a fault to have a causal connection $C$, and during the time it takes after a resolved downtime for components to heat back up to normal operation temperature $H$ (reheating).
            \begin{equation}
                D^i  = a \max_{t\not\in(F \cup H \cup C)}{\left(\mathrm{NOM}^i\left(\bm{T}_t\right)-T^i_t\right)^2}
            \end{equation}
            Here $a>1$ is the factor that determines the safety margin towards false alarms. A larger value reduces the risk of false alarms but also reduces the sensitivity to true anomalies. In this work, $a=1.2$ is used.
            The time between an anomaly and a failure can potentially be multiple months \cite{Vidal2023pmo}, but due to the dataset only spanning one year and one turbine, the training data is assumed to be free of anomalies after excluding all times during faults and reheating (i.e. $C$ is empty).
            It has been observed that reheating typically occurs within one hour after the turbine is operational again, and so all measurements up to one hour after turbine restart are grouped as reheating.
            For very long downtimes, it may take longer for the model to return to normal operation temperatures. Therefore, after a fault, all anomalies are classified as extended reheating until all the MSEs between NOM and measurement return below a second threshold $D_{ER}^i$, which is chosen smaller than $D^i$ to account for noise. Here, we choose
            \begin{align}
                D_{ER}^i = \max_{t\not\in(F \cup H\cup C)}{\left(\mathrm{NOM}^i\left(\bm{T}_t\right)-T^i_t\right)^2}
            \end{align}
            In addition to anomalies in the measurement, a test for sensor anomalies in the form of missing values is introduced through a counter. If a sensor is expected to measure at a time step, but no measurement is received for a number of consecutive values (here 10), a sensor anomaly is reported.
            The full algorithm for anomaly detection is 
            visualized in Figure \ref{fig:anom}. The fault diagnosis is presented in the following section.
            To test the algorithm, the confidence in an anomaly being connected to a fault needs to be quantified. To this end, the probability $\mathrm{p}()$ of the time difference between a random anomaly and a following fault of duration $d$ to be equal or smaller than the time difference $\Delta t$ of the detected anomaly and the following fault with duration $d$ is calculated.
            \begin{align}
                \mathrm{p}(\Delta t, d) &= \frac{\mathrm{n}(\Delta t, d)}{N} \label{eq:prob}\\
                \mathrm{n}(\Delta t, d) &= \sum_{i=1}^N \begin{cases}
                    1 ,& t_{\,\mathrm{F}_{i,d}} -t_i<\Delta t \\
                    0 ,& else
                \end{cases}
            \end{align}
            Here, $N$ is the total number of time steps where an anomaly could trigger, and $\mathrm{F}_{i,d}$ is the next fault after timestep $t_i$ with minimum duration $d$. A lower probability corresponds to a higher confidence that a detected anomaly is associated with the following fault.

        \subsection{Diagnostics}
        \label{ssec:m_diag}
            Each temperature signal is estimated by a separate NOM, but a detected anomaly in one temperature does not necessarily have to be caused by that temperature measurement. Instead, another abnormal input parameter could have caused a deviation of the NOM prediction from the measured value. To identify the signal responsible for the anomaly, the inputs most impacting the model output are identified using the Shapley additive explanations (SHAP) algorithm \cite{Lundberg2017aua}. The algorithm is based on game theory and can be used to identify the (input) feature importance for local and global model outputs.
            When an anomaly is detected in a signal, the local SHAP values are calculated. Three cases are considered:
            \begin{itemize}
                \item[1.] A single input signal dominates the output of the NOM with the anomaly and the signal is the same as the predicted signal.
                The NOM is corrupted by the input, the measurement doesn't behave as expected, or both. Either way, the anomaly is caused by that signal.
                \item[2.] Multiple signals contribute significantly to the output, but the contribution is not higher than in the training set during normal operation. The NOM performs as expected, but the measured value doesn't match the output signal. The anomaly is caused by the output signal.
                \item[3.] One or few signals contribute significantly more than expected from normal operation SHAP values. The anomaly is likely caused by a corrupted NOM input, which indicates an anomaly in those signals. An anomaly in the corresponding NOM is expected.
            \end{itemize}
            In cases 1 and 2, the anomaly can clearly be attributed to the output signal and is reported as such. In case 3, information from the other NOMs is required to pinpoint the anomaly. An anomaly is reported together with the unusually high feature contributions.
            To classify the cases, a signal $T_t^i$ is considered to dominate the output (case 1) if it constitutes more than 70\% of the total impact, that is 
            \begin{align}
            \left|\mathrm{SHAP}\left(\mathrm{NOM}^i\left(\bm{T}_t\right)\right)[i]\right| > D_{S1,t}^i
            \end{align}
            where the time-dependent thresholds $D^i_{S1,t}$ are calculated from 
            \begin{align}
            D_{S1,t}^i= 0.7 \sum_j\left|\mathrm{SHAP}\left(\mathrm{NOM}^i\left(\bm{T}_t\right)\right)[j]\right|
            \end{align}
            A signal $T^j$ is classified as contributing more than expected from the normal operation training data if 
            \begin{align}
                \left|\mathrm{SHAP}\left(\mathrm{NOM}^i\left(\bm{T}_t\right)\right)[j]\right| >D_{S2}^{ij}
            \end{align}
            with the time-independent thresholds $D_{S2}^{ij}$ defined as
            \begin{align}
                D_{S2}^{ij} = 1.2 \max_{t^\prime\not\in(F \cup H \cup C)}\left|\mathrm{SHAP}\left(\mathrm{NOM}^i\left(\bm{T}_{t^\prime}\right)\right)[j]\right|
            \end{align}

        \subsection{Digital Twin Integration}
\begin{figure}
                \centering
                \includegraphics[width=\linewidth]{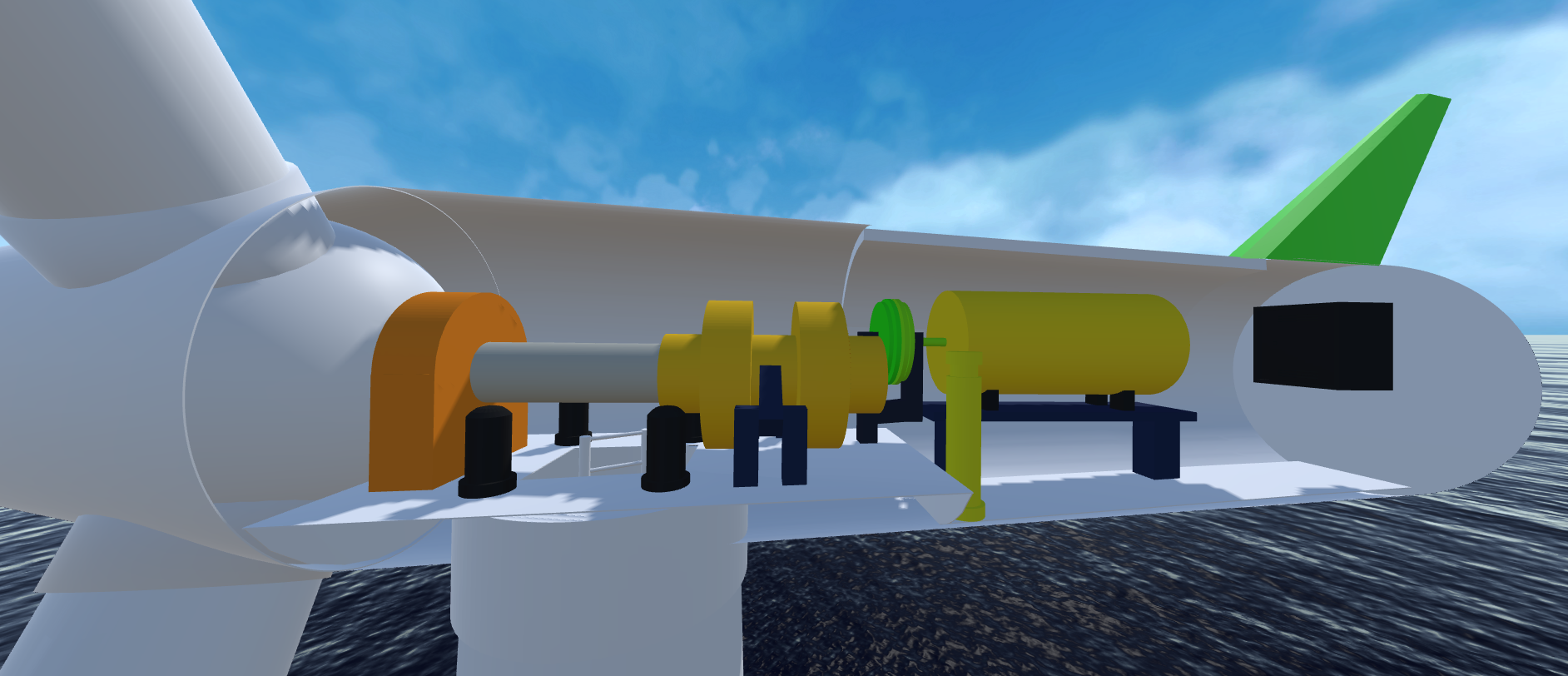}
                \caption{Digital twin virtual reality interface with components colored according to the measured temperature.}
                \label{fig:vr}
            \end{figure}

            The temperature data is visualized in the digital twin interface, which is built into the Unity engine.
            The interface is integrated as a cross-platform application for computers and virtual reality headsets. A detailed description of the digital twin interface can be found in~\cite{Stadtmann2023doa} and the data integration is described in~\cite{Stadtmann2024dif}.
            In addition to the graphical interface described in those articles, the temperature of components is visualized through colors to support diagnostics. A glimpse of the virtual reality integration is shown in Figure~\ref{fig:vr}.
            However, the turbine may not be supervised at all times. To that end, an alarm system is integrated that, if an alarm is triggered, sends a notification on mobile via SMS and messaging services containing the faulty farm, turbine, responsible component, signal type, and sensor, as well as a timestamp. 
            An example is shown in the Appendix in Figure \ref{fig:sms}.
        
    \section{Results and Discussion}
    \label{sec:Results and Discussion}
        The DNN NOMs are tested and integrated into the anomaly detection algorithm, the diagnosis approach is applied, and the results are presented here. The anomaly detection is repeated for the LSTM, and the results are compared with the DNN approach. Finally, diagnostic digital twins are discussed in a broader context of offshore engineering.
        \begin{figure}[t]
                \centering
                \includegraphics[width=\linewidth]{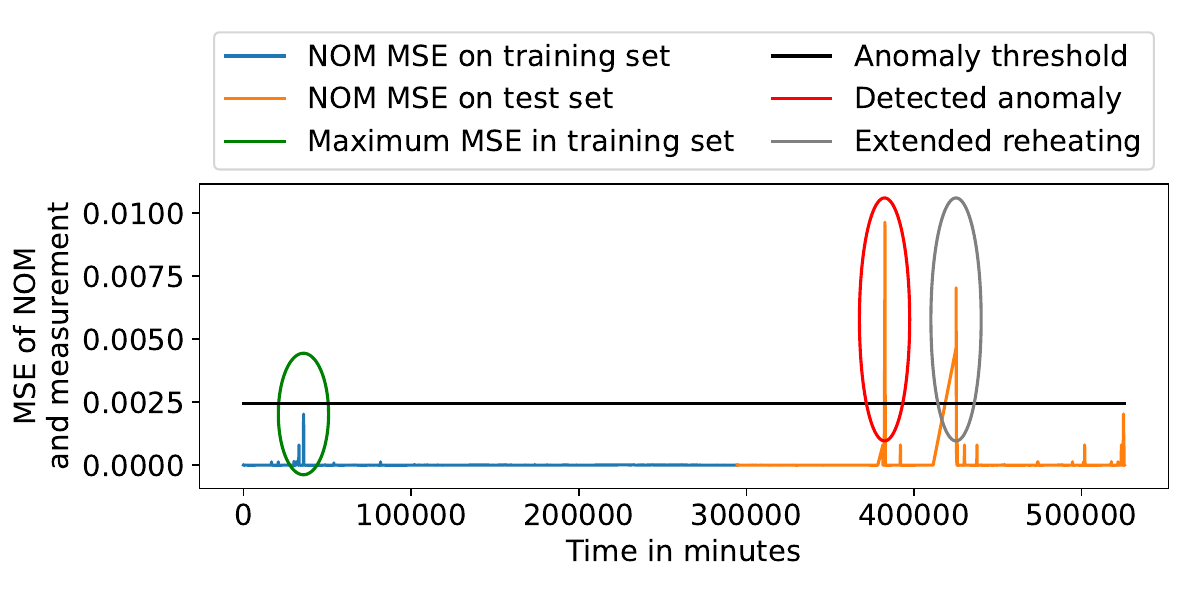}
                \caption{Detected anomaly in generator rotor temperature from mean squared error (MSE) between normal operation model (NOM) and measurement.}
                \label{fig:anomaly_result}
            \end{figure}
        \subsection{Dense Neural Network Normal Operation Model}
            First, the DNN NOM is evaluated on the training and testing data set. Several deviations between predicted and measured values are observed. However, almost all detected deviations are during and immediately after known downtimes, which can be explained by the previously mentioned cooling and reheating of components. By excluding known downtimes and reheating periods, the NOM has similar average deviations on training and test sets. 
            Despite the different seasons and therefore environment temperatures in training and test set, no seasonal bias is observed.
        
        \subsection{Anomaly Detection}
            
\begin{figure}[t]
                \centering
                \includegraphics[width=0.9\linewidth,trim={0 15 0 15}, clip]{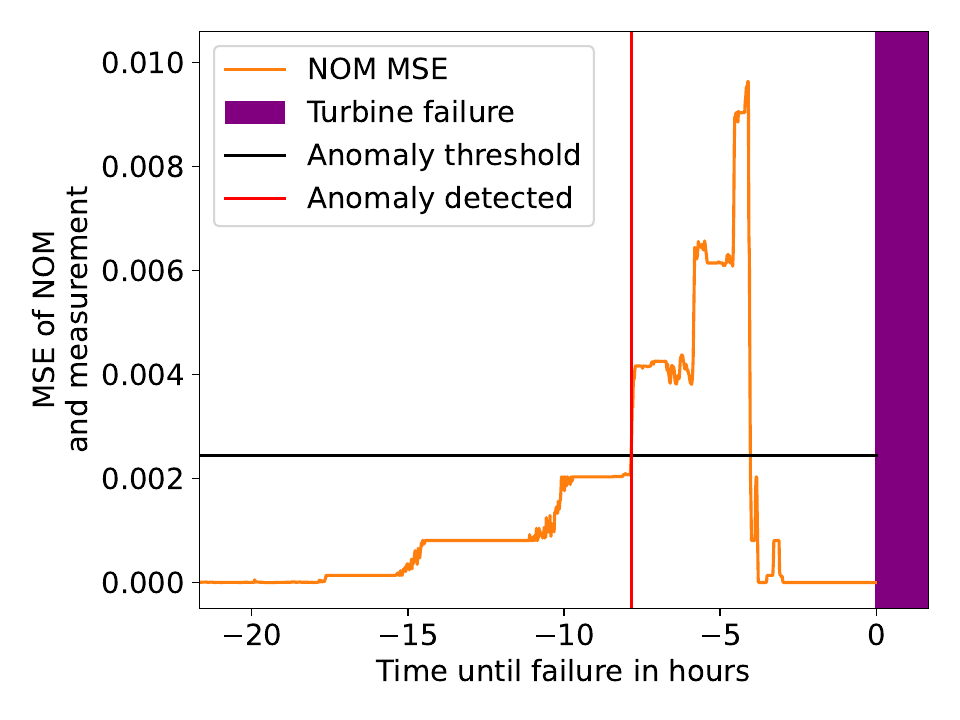}
                \caption{Zoom in on the detected anomaly in the generator rotor temperature.}
                \label{fig:anomaly_result_zoom}
            \end{figure}

\begin{figure}[t]
                \centering
                \includegraphics[width=\linewidth, trim={0 10 0 30}, clip]{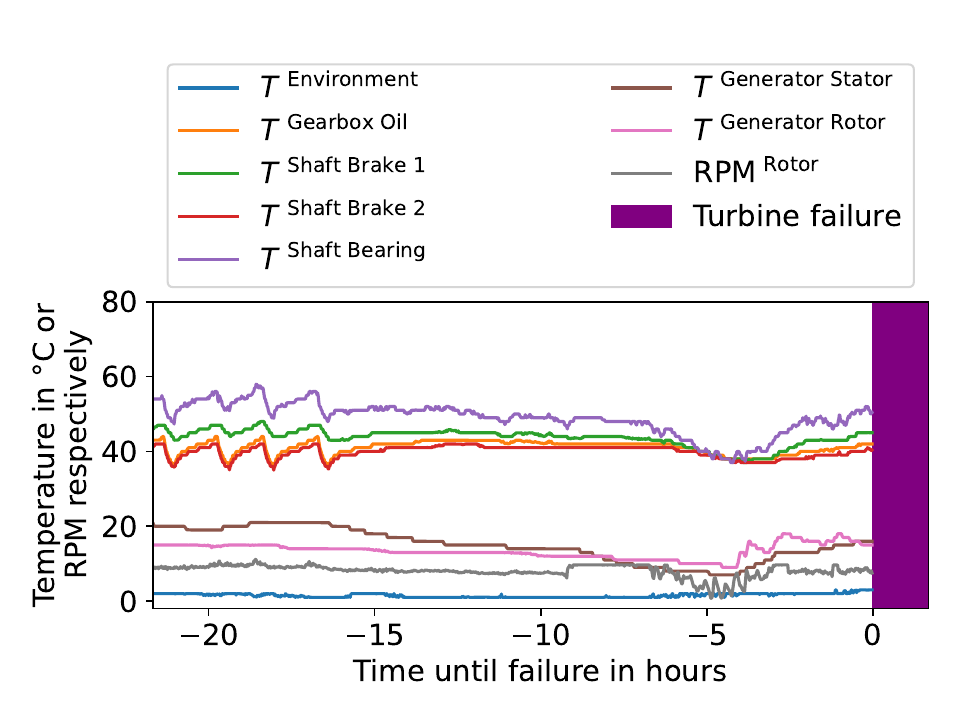}
                \caption{Zoom in on the data set around the detected anomaly.}
                \label{fig:anomaly_data_zoom}
            \end{figure}

            The MSEs between NOMs and measurements of the shaft main bearing temperature, the shaft breaks temperatures, and the generator oil temperature do not exceed the anomaly threshold. No false positives are detected.
            The generator rotor and the generator stator temperature MSE surpassed the anomaly threshold twice, one of which was filtered out by the algorithm as extended reheating.
            Both the generator rotor and stator temperatures detect the same anomaly, but the anomaly threshold is triggered 94 minutes earlier in the generator rotor temperature. The results of the anomaly detection applied to the DNN NOM results for the generator rotor temperature are shown in Figure~\ref{fig:anomaly_result}. An enlarged area around the anomaly is shown in Figure~\ref{fig:anomaly_result_zoom}.
            7.8 hours after the detected anomaly, the turbine stopped for 4.2 days.
            The probability of a detected anomaly occurring randomly equal or less than 7.8 hours before a fault of at least 4.2 days duration is $\mathrm{p}(7.8\,\mathrm{h},4.2\,\mathrm{d} )< 0.2\%$ (with $\mathrm{p}(\Delta t, d)$ as defined in Equation~\ref{eq:prob}). This strongly indicates that the anomaly has a causal connection to the fault.
            The endogenous and exogenous signals are plotted around the detected anomaly and leading up to the fault in Figure~\ref{fig:anomaly_data_zoom} to show how difficult it can be to identify an anomaly without a dedicated detection model.
            To ensure that the work is reproducible, the training process of the DNN NOMs has been repeated two additional times for a total of three evaluations, and the anomaly detection algorithm has been applied. The resulting NOMs differ only slightly from each other, and the differences do not affect the results of the anomaly detection algorithm. This robustness in training can be partially attributed to the fact that each NOM is selected as the best among three training repetitions as explained in Section~\ref{ssec:m_nom}.
            
        \subsection{Diagnosis}
\begin{figure}[t]
                \centering
                \includegraphics[width=\linewidth, trim={0 12 0 10}, clip]{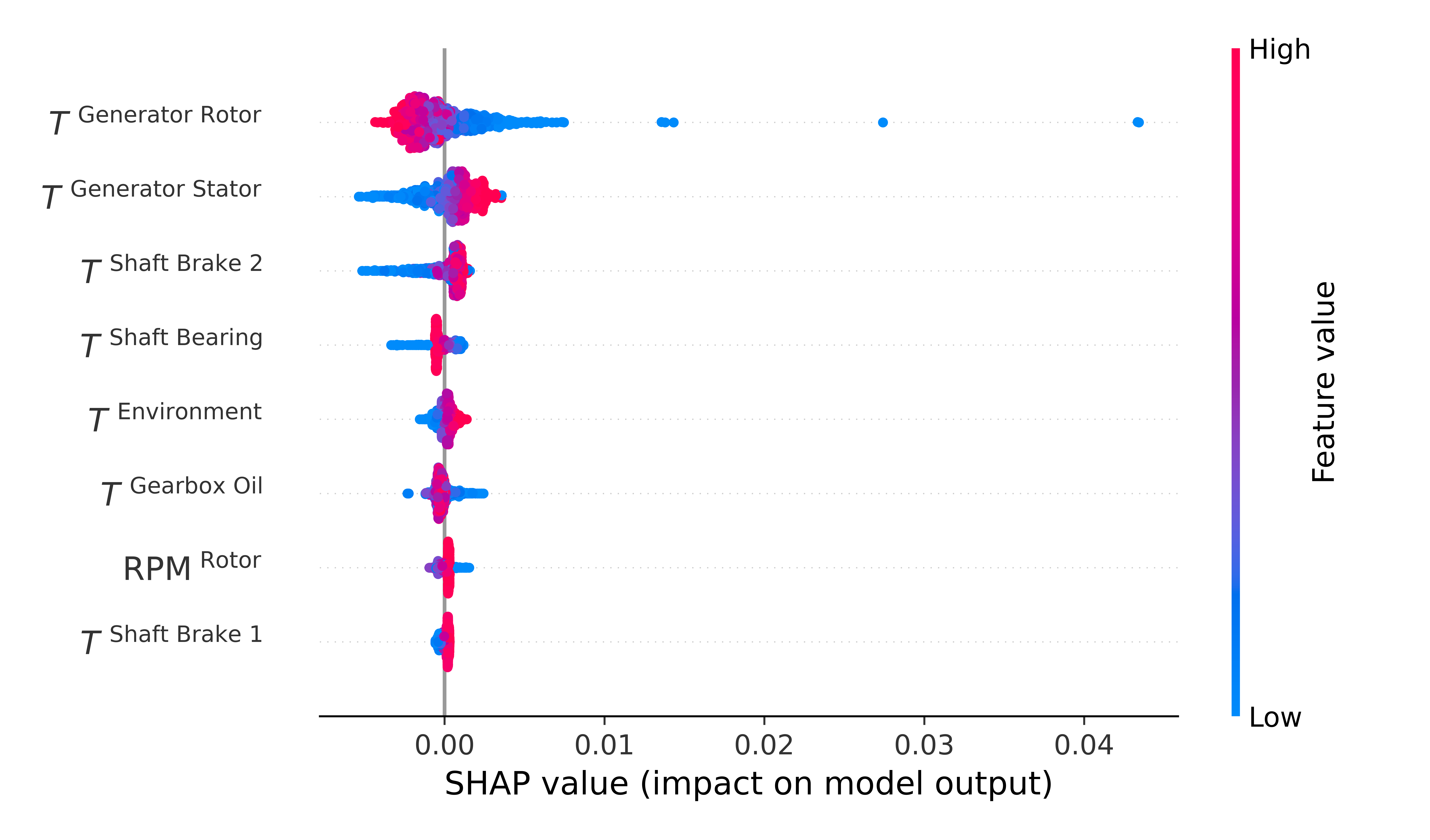}
                \caption{Shapley additive explanations (SHAP) values for generator rotor temperature normal operation model on 1000 samples from training data. Output depends on multiple inputs, and correlations are visible.}
                \label{fig:shap_train}
            \end{figure}

\begin{figure}[t]
                \centering
                \includegraphics[width=\linewidth,trim={0 12 0 10}, clip]{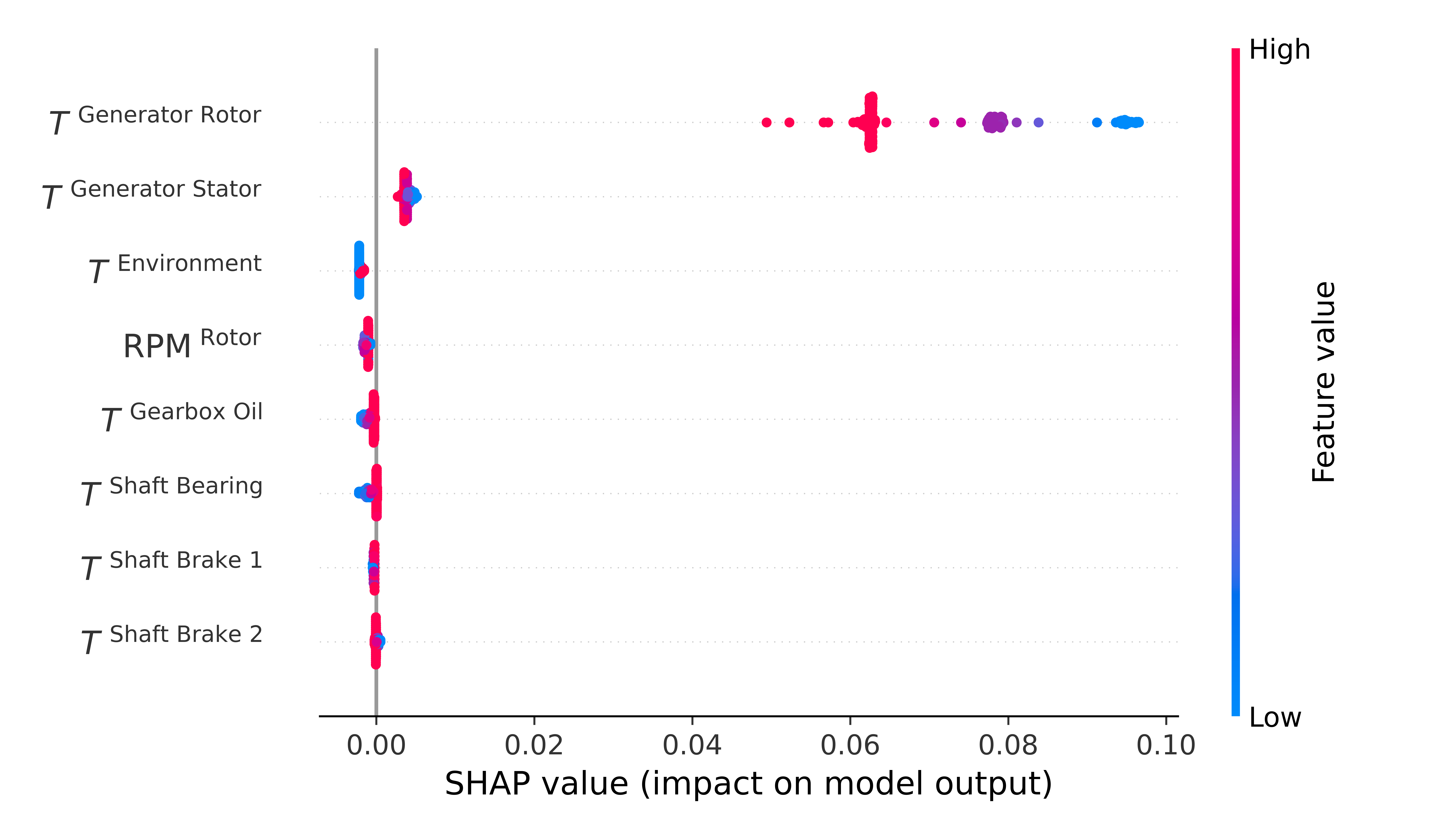}
                \caption{Shapley additive explanations (SHAP) values for generator rotor temperature normal operation model during anomaly. Output is dominated by generator rotor temperature input.}
                \label{fig:shap_anom}
            \end{figure}

            To diagnose the anomaly cause, SHAP is applied for each NOM on the training set. The SHAP values are plotted in Figure~\ref{fig:shap_train} for 1000 random samples from the training data. It is confirmed that while each NOM's output is affected strongest by the corresponding temperature input, the other inputs contribute to the output as they are correlated.
            To identify the anomaly's root cause, SHAP is applied in the abnormal interval to the NOMs where the anomaly is detected. The SHAP values are plotted in Figure~\ref{fig:shap_anom} for the samples that trigger an anomaly. It can be seen that during the anomaly, the SHAP values of the temperature input increase significantly over the other values, corresponding to case 1 in Section~\ref{ssec:m_diag}. This confirms that the anomaly is caused by the temperatures of the NOMs where the anomaly has been detected, that is by the generator rotor and stator. Therefore, the diagnosis suggests that the root cause of the anomaly lies with the generator.
            While here the SHAP values have only been used to confirm the fault cause, the anomaly did manifest directly in the SHAP values. The SHAP values may provide the basis for the anomaly detection itself, but further research with more extensive data sets is required here.

        \subsection{Long Short-Term Memory Normal Operation Model}
        
            The anomaly detection method is repeated with the LSTM model on the same data set. The results can be found in the appendix in Figure~\ref{fig:lstm_anom}. A spike in the NOM MSE is observed where the DNN NOM has detected the anomaly, but the spike is not strong enough to trigger an anomaly alarm. A reduction of the threshold could be considered, but this increases the risk of false alarms. However, even with a reduced threshold, the MSE becomes significant only after the DNN already reports the anomaly.
            Two anomalies are detected at other times, but no turbine downtimes lasting longer than 1 hour have been recorded within weeks after these anomalies. However, it should be noted that up to several months can pass between a detected anomaly and a fault~\cite{Vidal2023pmo}. Therefore, it cannot be ruled out that the anomalies have a causal connection to faults occurring weeks or months later.
            Shorter downtimes of less than 30 minutes and 15 minutes occur within 18 hours and 2.3 days respectively. The probabilities for anomalies showing up this close to a fault of at least the corresponding length of downtime are $\mathrm{p}(18\,\mathrm{h},30\,\mathrm{min}) = 16\%$ and $\mathrm{p}(2.3\,\mathrm{d},15\,\mathrm{min})  = 37\%$ respectively (Equation~\ref{eq:prob}). However, these short downtimes only have minor contributions to the total downtime, while the anomaly detected by the DNN and confirmed with much higher certainty has a substantial contribution but was not picked up by the anomaly detection algorithm as clearly using the LSTM NOM.
            Similar to the DNN NOM, the analysis is repeated twice, but in contrast to the DNN, the changes in the LSTM NOM are strong enough to propagate into the anomaly detection process. In both additional cases, different anomalies are detected, none of which can be connected to a fault with confidence.
            It could be considered to use an ensemble of LSTMs for a more reproducible result. However, for a more detailed test of the algorithms and comparison of DNNs and LSTMs, a significantly larger data set with several turbines and multiple years of data is required to include more faults and increase the potential causal horizon in the data set. Additionally, more detailed logs on the fault source and maintenance work would give certainty on the verification of the diagnosis process.

        \subsection{Diagnostic Digital Twins in Offshore Engineering}
            Diagnostic digital twins have potential in many diverse industries. However, they are especially valuable for offshore engineering applications, where assets are remote, response times long, manual inspections difficult or even impossible, and repairs are expensive and time-sensitive. 
            Whether monitoring rotating equipment with temperature, vibration, or acoustic emission signals, or static components with strain and stress sensors, the potential of diagnostic digital twins increases with decreasing sensor prices, increasing connectivity and computing power.
            However, while in some industries anomaly detection and component damage monitoring have been actively used for years, utilizing the increased potential of anomaly detection from huge amounts of data streams with various origins through physics-based models requires a large workforce with domain expertise. Unsupervised machine learning methods have the potential to solve this bottleneck as they can be applied with little to no domain knowledge.
            However, a lack of trust from the industry has been observed especially in data-driven methods, but also in physics-based methods due to a lack of model interpretability and uncertainty estimates. A missed anomaly may have severe consequences, and the cost of responding to false positives can be substantial.
            More user cases on diagnostic digital twins are required to the point where sufficient trust is built for the industry to start adopting and scaling the technology. The process may be sped up by running anomaly detection in real-time without acting on a detected anomaly yet. When sufficiently many faults have been predicted without false positives, a ``told-you-so" effect may quickly build enough trust to start relying on the methods. Currently, this approach is limited by a lack of access to real-time data in academia and the cost of expertise to integrate approaches in the industry, but a closer collaboration between industry and academia could close that gap in the near future.

    \section{Conclusion and Future Work} \label{sec:Conclusion}
        Digital twins of physical assets bring many benefits throughout the asset lifetime, and especially diagnostic digital twins have the potential to improve asset lifetime and maintenance strategies across many industries. 
        In this article, the concept of digital twins, capability levels, and diagnostic digital twins in particular have been explained. A diagnostic digital twin has been demonstrated for an existing floating offshore wind turbine using a neural-network-based normal operation model with feature importance explanation for anomaly detection and diagnosis in temperature signals. In contrast to most existing research on floating offshore anomaly detection, which relies on turbine simulations~\cite{Wang2023ddf,Li2020rao,Ghane2018cmo}, this work uses data from an operational floating wind turbine. 
        It was found that the small dense neural network trained unsupervised was able to detect a temperature anomaly that had not been identified in the commercial system of the asset and lead to an asset fault with multiple days of downtime 7.8 hours after the anomaly was first detected. 
        The sensor signals responsible for the anomaly were identified using shapely additive explanations (SHAP).
        The model was compared to a long short-term memory model, which was not able to clearly identify the anomaly.
        Training the model requires no explicit domain knowledge, and the model setup can be generalized to any other set of temperature signals or assets by retraining it on a corresponding data set.
        The anomaly detection algorithm has been interfaced with messaging services to send alarm notifications, and the integration into a virtual reality interface for intuitive fault identification has been outlined.
        However, the validation of the model has been limited to one year of historical data from a single turbine, which limits the validation of the NOMs. The analysis can be extended in future research as more data from floating offshore wind turbines becomes available for academic research.
        In general, more academic user cases on diagnostic digital twins are required to build sufficient trust for the industry to adopt and scale diagnostic digital twins. The process could be sped up by integrating the methods with real-time data in collaboration between academia and industry to build trust in the system and accelerate industrial integration.
    
    
    \section*{Acknowledgments}
    This publication has been prepared as part of NorthWind (Norwegian Research Centre on Wind Energy) co-financed by the Research Council of Norway (project code 321954), industry, and research partners. Read more at \url{www.northwindresearch.no}.
    Data for the project realization was provided by SINTEF Energy Research.
    

\bibliographystyle{asmeconf}
\appendix
\section{Additional figures}

\begin{figure}[]
    \centering
    \includegraphics[width=\linewidth]{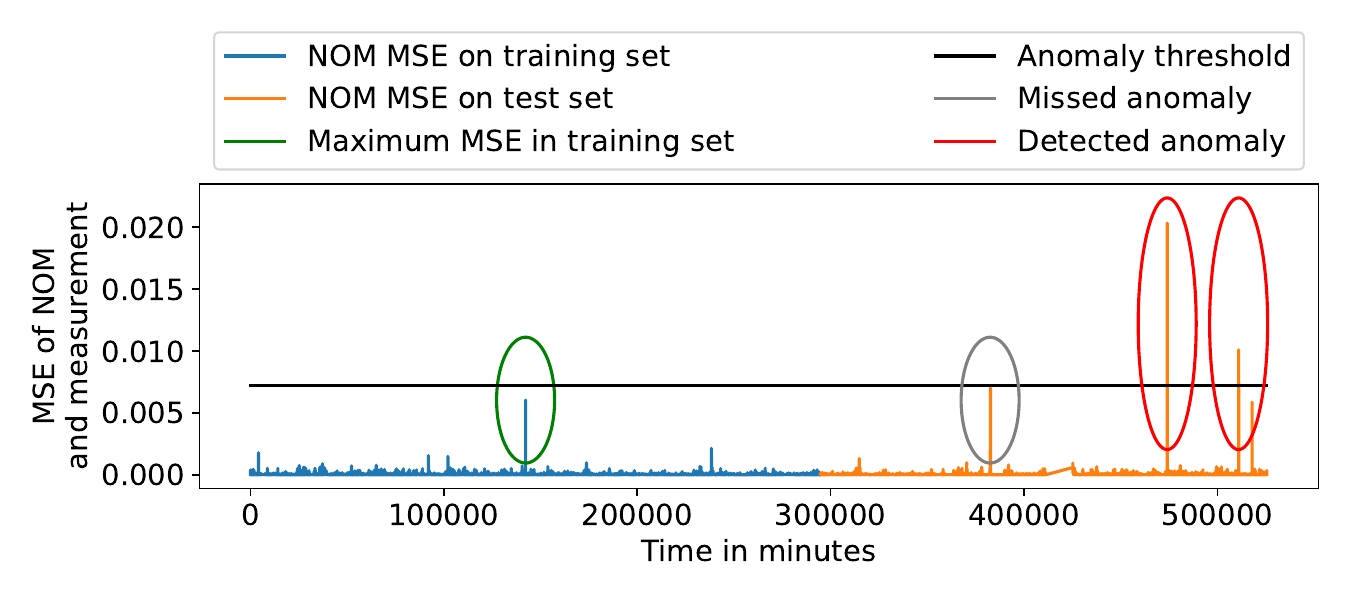}
    \caption{Mean squared error (MSE) between long short-term neural network (LSTM) normal operation model (NOM) and measurement for generator rotor temperature.
    }
    \label{fig:lstm_anom}
\end{figure}

\begin{figure}[]
    \centering
    \includegraphics[width=0.6\linewidth]{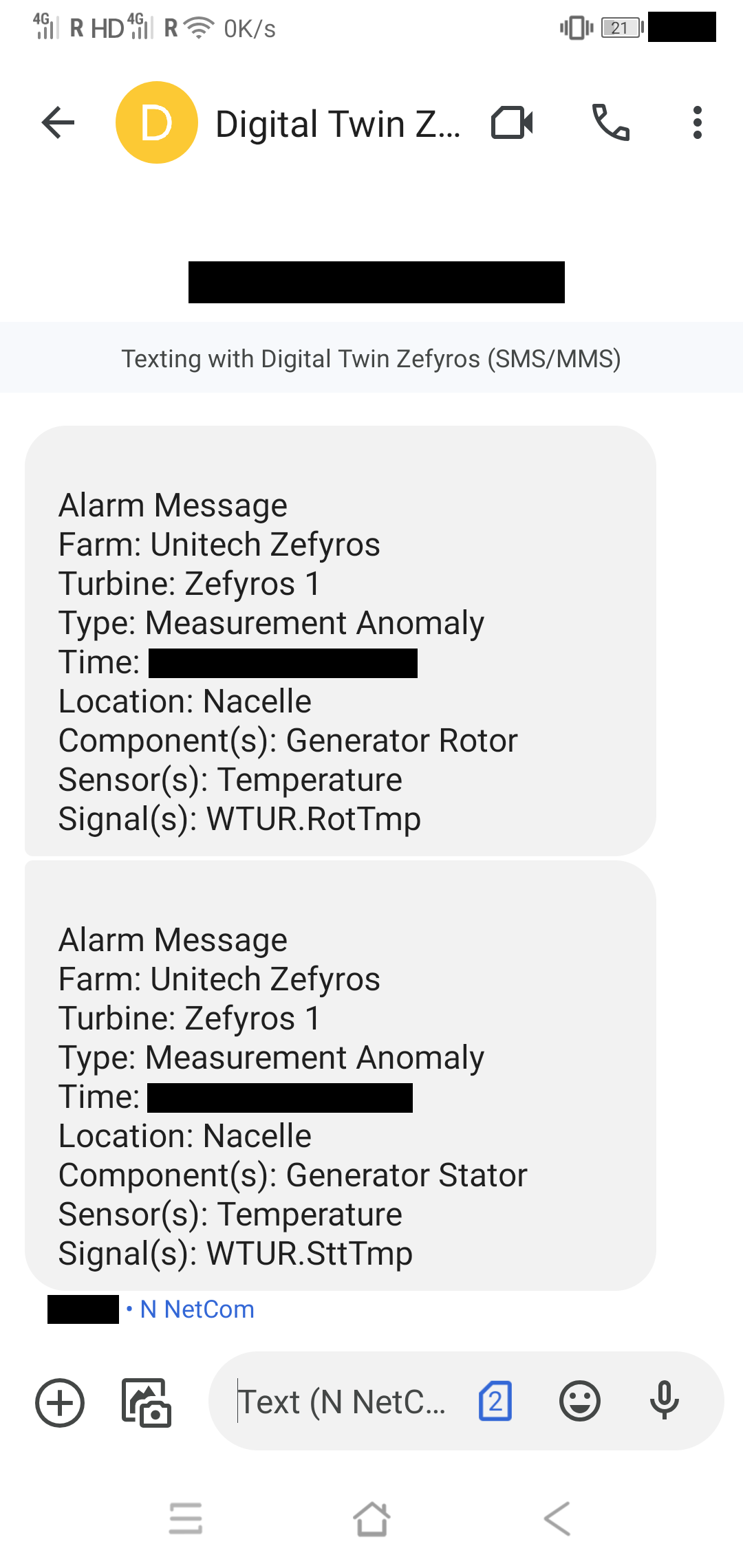}
    \caption{An alarm message sent via SMS notifies the turbine operator of the detected fault.}
    \label{fig:sms}
\end{figure}
	
\end{document}